\documentclass[conference,12pt]{IEEEtran}

\IEEEoverridecommandlockouts
\usepackage{cite}
\usepackage{graphicx}
\usepackage{amsmath,amssymb,amsfonts}
\usepackage{algorithmic}
\usepackage{graphicx}
\usepackage{textcomp}
\usepackage{xcolor}
\usepackage{mathtools}
\usepackage{balance}
\begin{document}

\title{CoachAI: A Project for Microscopic Badminton Match Data Collection and Tactical Analysis}

\author{
{Tzu-Han Hsu}$^{\dagger}$
{~ Ching-Hsuan Chen}$^{\dagger}$
{~ Nyan Ping Ju}$^{\dagger}$
{~ Ts\`i-U\'i \.Ik}$^{\dagger}$$^{\ast}$
{~ Wen-Chih Peng}$^{\dagger}$ \\
{~ Chih-Chuan Wang}$^{\ddagger}$\thanks{$^{\ddagger}$Office of Physical Education, National Chiao Tung University, Hsinchu City 30010, Taiwan.}
{~ Yu-Shuen Wang}$^{\dagger}$
{~ Yuan-Hsiang Lin}$^{\diamond}$\thanks{$^{\diamond}$Department of Electronic and Computer Engineering, National Taiwan University of Science and Technology, Taipei City 106, Taiwan.}
{~ Yu-Chee Tseng}$^{\dagger}$ \\
{~ Jiun-Long Huang}$^{\dagger}$
{~ Yu-Tai Ching}$^{\dagger}$ \\
$^{\dagger}$Department of Computer Science, College of Computer Science \\
National Chiao Tung University \\
1001 University Road, Hsinchu City 30010, Taiwan \\
$^{\ast}$Email: cwyi@nctu.edu.tw}

\maketitle

\begin{abstract}
Computer vision based object tracking has been used to annotate and augment sports video. For sports learning and training, video replay is often used in post-match review and training review for tactical analysis and movement analysis. For automatically and systematically competition data collection and tactical analysis, a project called CoachAI has been supported by the Ministry of Science and Technology, Taiwan. The proposed project also includes research of data visualization, connected training auxiliary devices, and data warehouse. Deep learning techniques will be used to develop video-based real-time microscopic competition data collection based on broadcast competition video. Machine learning techniques will be used to develop tactical analysis. To reveal data in more understandable forms and to help in pre-match training, AR/VR techniques will be used to visualize data, tactics, and so on. In addition, training auxiliary devices including smart badminton rackets and connected serving machines will be developed based on the IoT technology to further utilize competition data and tactical data and boost training efficiency. Especially, the connected serving machines will be developed to perform specified tactics and to interact with players in their training.
\end{abstract}


\section{Introduction}

In the high-rank badminton competitions, the outstanding performance of athletes depends on not only the persistent and solid training but also the mastery of the opponents' superiority, inferiority, and even tactics and emotion in the games. Hence, supporting systems for real-time tactical information collection and analysis are critical and significant for top athletes to win their games. Possible services during matches may include monitoring of ball type usage and ball accuracy, evaluating players' reaction agility and footwork techniques, alerting happening of continuous loss and analyzing the causes, and detecting modes of defensive or offensive playing. According to the information, coaches can advise players to take corresponding tactics. Furthermore, in the post game review, such a system can provides accurate gaming data for the discussion. Therefore, automatic competition data collection and tactical analysis are turnkey technologies for professional athletes.

Videos can be considered as logs of visual sensors and retain a large amount of information. In most cases, tactical data were manually labeled from video, but this approach is time-consuming and laborious. In the last few years, smartphone apps have been used to assist on-site labeling. Due to the fast rhythm of the games, mistakes happen from time to time. It would be efficient to extract competition data from video based on computer vision techniques, and real-time tactical analysis incorporated with big data analysis could be realized. This method not only reduces manual errors, but also improve efficiency and accuracy. Moreover, beyond macroscopic data such as ball types, computer vision can provide microscopic data such as shuttlecock trajectories and player postures frame by frame. Furthermore, AR/VR can help coaches and athletes to easily catch the key information and utilize the information in their training.

The Hawk-eye \cite{Hawk-Eye} is a multiple cameras system that can calculate 3D ball trajectories and has been widely used in various international competitions such as football, cricket, tennis, etc. to assist umpires in the judgments of controversial balls. In NBA, professional companies deploy high-resolution cameras in arenas incorporated with offline image processing for the tracking of the ball and players. However, these systems are expensive and accompanied high operating cost. In the study of shuttlecock trajectory prediction \cite{Badminton:2016:Waghmare}, 2D laser scanners are used to locate the badminton shuttlecock in the real world environment. In addition, shuttlecock speed and orientation, etc. can be known. However, the equipment is not popular and lead to limited application. In \cite{Badminton:2017:Chu}, computer vision incorporated with machine learning are applied to detect the court and players and classify ball types. The data of player positions and ball types are used in the classification of tactics. Most works mentioned require the assistance of experienced coaches. That hinders the development of automatic data collection systems.

\begin{figure}[ht]
    \centering
    \includegraphics[width=0.48\textwidth]{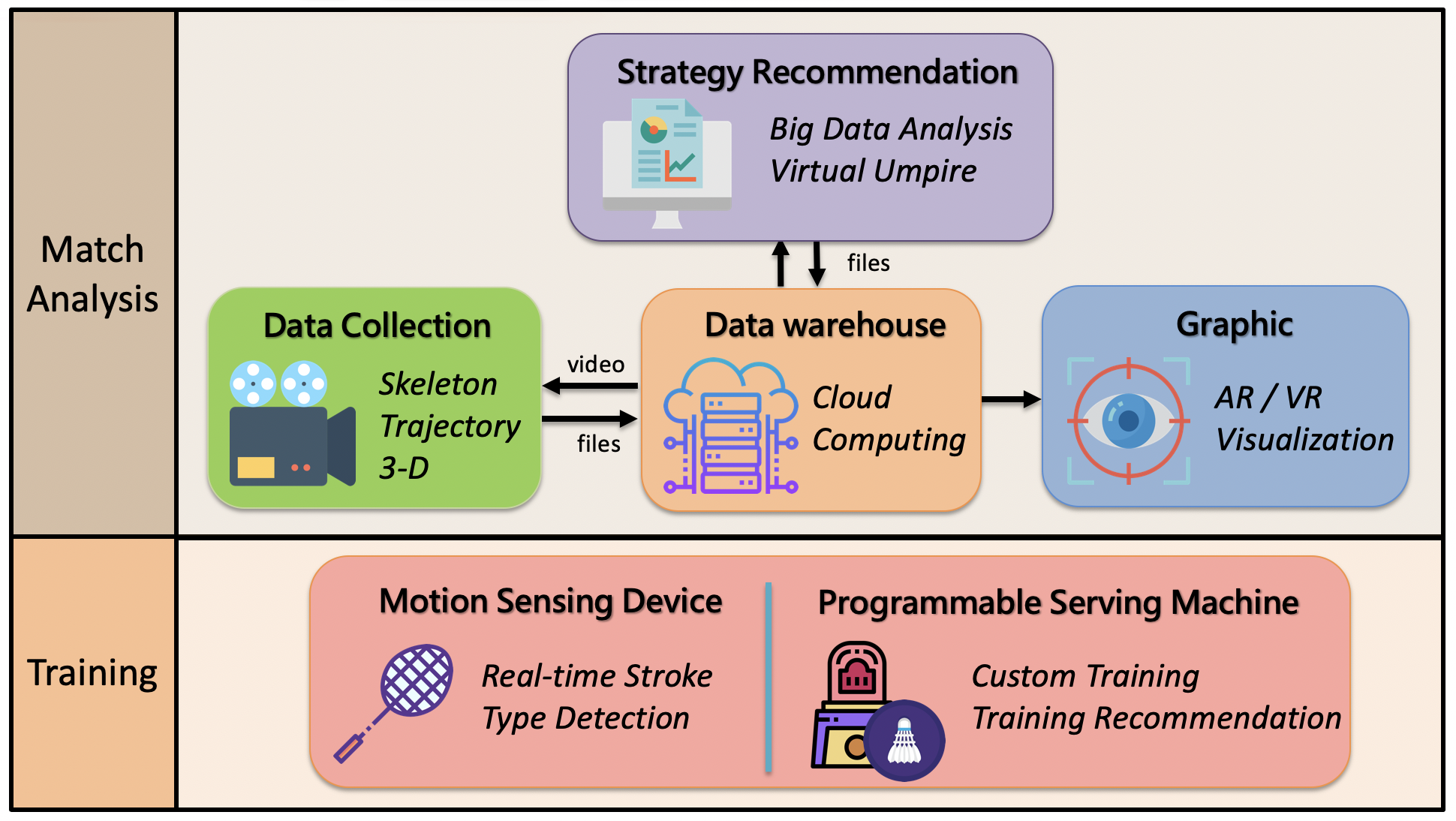}
    \caption{The scope of CoachAI.}
    \label{fig:project}
\end{figure}

In this paper, a project for badminton competition data collection and tactical analysis is introduced. Fig. \ref{fig:project} depicts the scope of this project, a mobile computing and cloud service based solution. The turnkey techniques include deep learning video analysis, machine learning and big data tactical analysis, wearable sensor and IoT technology, and AR/VR visualization. The data collection sub-project utilizes broadcast videos from public medias such as the BadmintonWorld.TV channel on YouTube or videos taken by mobile devices or consumer cameras to efficiently collect microscopic competition data by deep learning video analysis. TrackNet \cite{TrackNet:2018:Huang} is used to track the shuttlecock, YOLOv3 \cite{YOLOv3:2018:Redmon} is used to locate the players, and OpenPose is used to detect the skeletons of the players. The tactics sub-project develops ball type classification and spatial and temporal tactical analysis. Real-time court-side feedback will be possible for coaching via mobile devices. The visualization sub-project provides graphic user interfaces and data representation for easy understanding and speedy knowledge discovery. The data warehouse sub-project provides the data exchange platform and keeps all microscopic and macroscopic competition data for further analysis. Most of the services will be hosted on cloud platforms. In addition, an IoT-based serving machine system that can be programmed to perform specified tactics will be developed to interact with players in their training.

Overall speaking, we are going to build the capability of microscopic data collection and tactical analysis of badminton competitions. In addition, visualization techniques such AR/VR will be used to help data presentation and data mining, and connected serving machines and smart rackets will be developed to boost training efficiency. Finally, back-end servers will be built to host cloud services and data warehouse. The rest of this paper is organized as below. Section \ref{sec:CoachAI} presents the architecture of the proposed system in the CoachAI project. In Section \ref{sec:TrackNet}, the deep learning network for shuttlecock trajectory tracking is described. Section \ref{sec:Posture} introduces the deep learning networks for player positioning and skeleton detection. Section \ref{sec:Implementation} provides current implementation status. The conclusion and future works are given in Section \ref{sec:Conclusion}.

\section{Overview of the CoachAI Project}
\label{sec:CoachAI}

The scope of the CoachAI project is illustrated in Fig.~\ref{fig:project}. The data collection sub-project, Sub-project 1, is going to build the capability of automatic microscopic data collection based on deep learning technology. The videos of top-rank professional badminton competitions, that are available on public domain such as All England Open Badminton Championships on YouTube, are the subjects to be analyzed. The frame-level microscope data include the position of the shuttlecock, the locations and skeletons of the players. For each frame, a deep learning network, called TrackNet, that can locate high-speed and tiny objects from ordinary broadcast video is adopted to position the flying shuttlecock; and the well known YOLOv3 and OpenPose packages are fine-tuned to segment players and detect the skeletons of players, respectively. More macroscopic data, such as shuttlecock speeds, stroke times, ball types and footworks, can be available based on the microscope data.

The tactical analysis sub-project, Sub-project 2, is going to build the capability of tactical analysis and recommendation based on machine learning techniques and big data analysis. The tactical information not only will be provided for athletes and coaches' reference but also can be further used to program the connected serving machines to boost the training efficiency. Classifiers will be designed to identify stroke times, recognize ball types and proactive/active modes of playing, and predict the drop point of the shuttlecock. Statistic analysis can be applied to analyze the footwork patterns and to find the reasons of continuous loss. Spatial analysis can help to understand the ability of ball control and to find ball usage patterns. Temporal analysis can be used to estimate physical declination and to detect tactical change. During the matches, the capability of real-time tactical analysis is important in order to provide tactical information for coaches and players.

The visualization sub-project, Subproject 3, provides graphic user interfaces, image processing functionality, and AR/VR capability. By the help of Subproject 3, the data obtained by Subproject 1 and Subproject 2 can be presented in the ways of easy understanding. That not only helps athletes and coaches to quickly catch the insight of the data but also benefits data mining processes. Furthermore, virtual reality (VR) and augmented reality (AR) techniques can help athletes familiar with opponents' ball type usage, footwork, and tactics in immersed ways. That can strengthen psychological quality of players.

\begin{figure}[ht]
    \centering
    \includegraphics[width=0.48\textwidth]{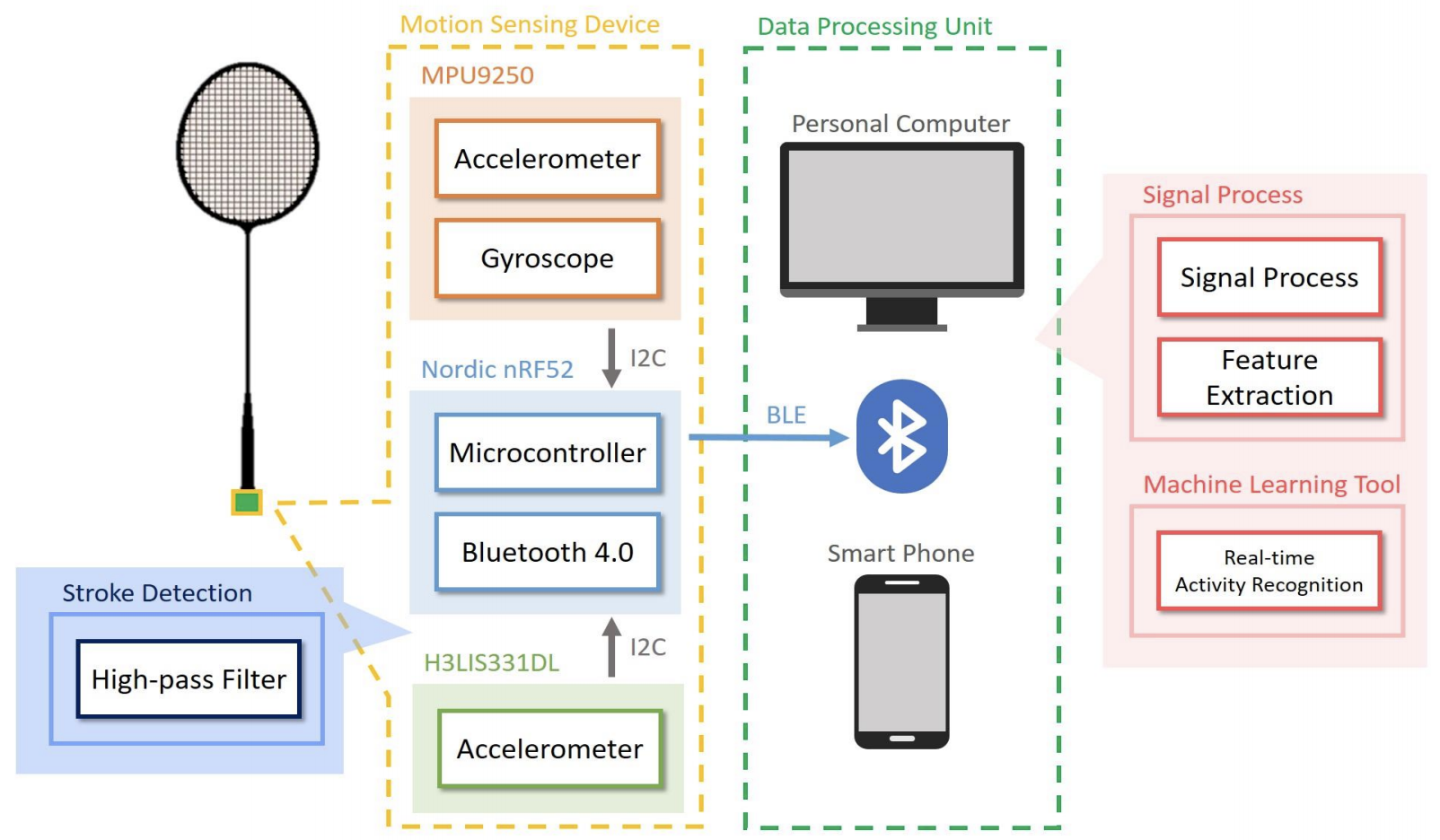}
    \caption{The architecture of Smart rackets.}
    \label{fig:system}
\end{figure}

In addition to data collect and tactical analysis, training auxiliary devices can directly benefit training efficiency. The training assistant sub-project, Subproject 4, is going to develop a real-time training auxiliary system including smart badminton rackets and connected serving machines. Besides the computer vision based ball type classification, smart rackets are designed to recognize stroke types by utilizing wearable IMU sensors installed at the bottom of the racket handle. The wearable sensors are composed of MPU9250 and Nordic nRF52 chips. The MPU9250 chip has a 3-axis accelerometer and a 3-axis gyroscope, and the nRF52 chip supports the Bluetooth 4.0 protocol stack. An app is developed to collect the IMU data and classify strokes. The system architecture of the smart rackets is illustrated in Fig.~\ref{fig:system}, and the app UI is depicted in Fig.~\ref{fig:interface}. In addition, IoT-based connected serving machines provide APIs for programmable online operation. Therefore, the serving machines can simulate opponent's ball usage patterns to reinforce pre-match training.

The data warehouse sub-project, Subproject 5, provides database and cloud computing services. It stores a large amount of video data, trajectory data and sensor data to support the development of other sub-projects, especially for tactical analysis. Besides, it provides cloud computing services such as deep learning, big data analysis or other complex operations. Parallel computing will be used to accelerate calculation, and docker technology will be introduced to fully utilize GPU resources.

We cooperate with the school badminton team of NCTU as well as top-ranked athletes in Taiwan through the coordination of the Chinese Taipei Badminton Association. To help national athletes to improve their performance, the developed techniques will be used to collect top 16 athletes in the world ranking in the future. At the current stage, the NCTU badminton team members provide their professional knowledge on data labeling and data representation. They also provide professional consultation on tactical analysis and data presentation. The prototype system will be assessed by the school team to take feedback from the perspective of professionals. The value of the prototype system will be finally evaluated by the national athletes. We expect to provide players with a big data analysis system that combines real-time strategy analysis and training-assisted devices. By Strategy Recommendation system, players can adjust their tactic during the game instantly. By real-time training auxiliary devices, players can adjust the training content to improve the training efficiency. The proposed system can enable players to have outstanding performance and provide players a better training environment.

\section{Shuttlecock Trajectory Tracking}
\label{sec:TrackNet}

It is challenging to track small objects from low resolution or low quality video. Even worse is that for sports like tennis, badminton, and baseball, due to the small size and high flying speed of the balls, the images of the balls in the video are typically blurred, and the ball recognition and positioning become harder. In addition, occlusion is another issue that may hinder recognition algorithms. Currently, commercial solutions in the market usually rely on high resolution and high frame rate video. However, the hardware investment and operation cost are also high.

\begin{figure}[ht]
    \centering
    \includegraphics[width=0.48\textwidth]{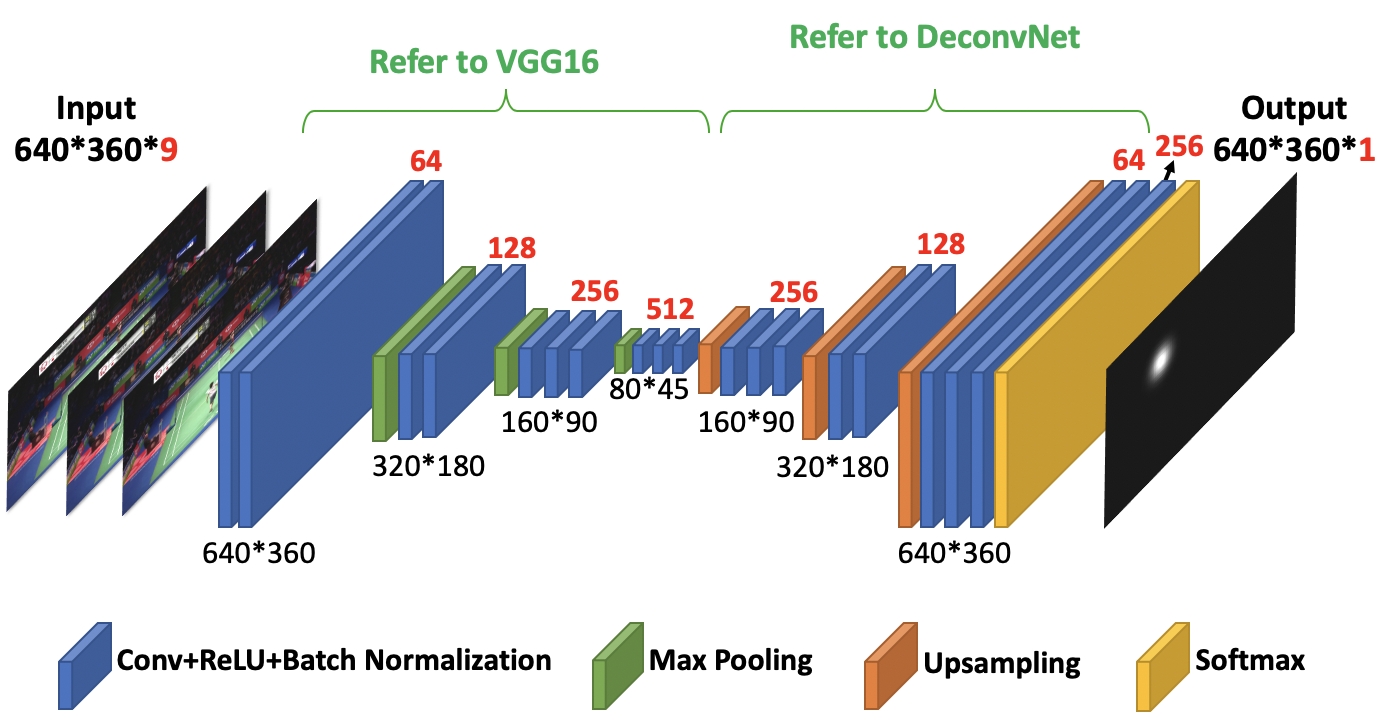}
    \caption{The architecture of TrackNet}
    \label{fig:TrackNet}
\end{figure}

A convolutional neural network (CNN) based framework, called TrackNet \cite{TrackNet:2018:Huang}, has been proposed to take consecutive frames together to generate a heatmap for ball detection. Fig. \ref{fig:TrackNet} depicts the framework of TrackNet. In the figure, TrackNet is designed to process three consecutive frames at once, and to output a ball detection heatmap corresponding to the last frame. Since three frames are input together, not only the appearance but also the trajectory of the shuttlecock could be cues for ball detection and positioning. In this work, we adpoted TrackNet to track the high speed shuttlecock from broadcast videos or those token by consumer mobile devices such as smartphones.

For training and testing purposes, a dataset based on videos available on Youtube with resolution $1280\times 720$ and then downsampled to $640\times 480$ has been prepared. TrackNet is trained to generate a probability-like detection heatmap that has the same size as the input images. The first 13 layers referring to the design of VGG16 \cite{VGG:2014:Simonyan} whose input data array has dimension $640\times 480$ is for the purpose of feature extraction, and the 14th to 24th layers referring to DeconvNet \cite{DeconvNet:2015:Noh} follows to perform upsampling. At the end, a ball detection heatmap is generated by applying pixel-wise softmax. The ground truth of a heatmap is an scaled 2D Gaussian distribution located at the head of the shuttlecock. The ball center is available in the dataset, and the variance of the Gaussian distribution refers to the size of shuttlecock images. The formula is showed in Eq. (\ref{eq:gaussian}).
\begin{equation}
G(x,y) = \lfloor(\frac{1}{2\pi\sigma^2}e^{-\frac{(x-x_0)^2+(y-y_0)^2}{2\sigma^2}})(2\pi\sigma^2\cdot255)\rfloor
\label{eq:gaussian}
\end{equation}
The first part is a Gaussian distribution centered at $(x_0, y_0)$ with a variance $\sigma^2$. The second part scales the value to the range $[0, 255]$. Since the average radius of the shuttlecock in the images is about 5 pixels, we set $\sigma^2 = 10$. Fig.3 depicts an example of the heatmap function.
\begin{figure}[ht]
    \centering
    \includegraphics{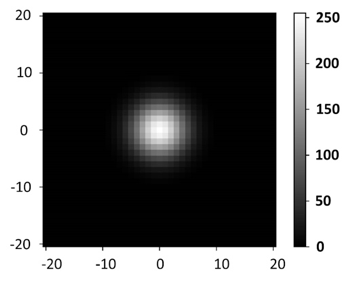}
    \caption{Heatmap example.}
    \label{fig:heatmap}
\end{figure}

The detection heatmap is not directly used in the calculation of the loss function, and therefore is not considered as part of the deep convolution network. The last second layer that is the one right before the softmax operation has size $640\times 480$ and depth $256$. The size is the same as the input image size, and the depth is corresponding to the grayscale values from 0 to 255. The pixel-wise one hot encode is used to encode heatmap into the output matrix. The pixel with the largest value on the heatmap is considered as the location of the shuttlecock. In the training phase, the cross-entropy function is used to calculate the loss function of $P(i,j,k)$. $P(i,j,k)$ denotes the pixel-wise probability of gray-scale value $k$ at position $(i,j)$, and $Q(i,j,k)$ is the corresponding ground truth. Let $H_{Q}(P)$ denote the loss function. Then,
\begin{equation}
H_{Q}(P)=-\sum_{i,j,k}{Q(i,j,k)logP(i,j,k)}
\label{eq:loss}
\end{equation}

After all, if a detection heatmap is given, the position of the shuttlecock can be decided as follows. First, the heatmap is converted into a black-white binary map by a threshold. If the value of a pixel on the heatmap is greater than the threshold, the output of the pixel is set to 255. Otherwise, the output is set to 0. Then, a circle finding algorithm named Hough Gradient Method is applied to find the center of each spot. If exactly one circle is identified, the center of the circle is reported as the location of a shuttlecock. In other cases, the image is considered as without a shuttlecock.

\section{Athlete Posture Detection}
\label{sec:Posture}

In order to evaluate the movements of players, the posture data are important. The collection of posture data is divided into two stages, the bounding box detection and the skeleton detection. YOLOv3 \cite{YOLOv3:2018:Redmon} is used in the player bounding box detection, and OpenPose \cite{OpenPose:2018:Cao} is used in the player skeleton detection.

Object detection needed in many computer vision applications is one of the most early studied problems in deep learning. The R-CNN family \cite{R-CNN:2014:Girshick,FastR-CNN:2015:Girshick,FasterR-CNN:2017:Ren} and YOLO family \cite{YOLO:2016:Redmon} are two main frameworks in object detection. R-CNN-like networks first generate many region proposals each of which might enclose an object, and then detect and classify the object in each region proposal. To speed up the processing speed, YOLO-like networks apply end-to-end philosophy. A set of pre-defined grid-based regions will be examined to estimate the probability of existing some kinds of objects. YOLO's are considered as good choices for real-time applications. YOLOv3 is currently one of the widely used object detection networks. It is worth mentioned that at resolution $320\times 240$, YOLOv3 runs in 22 ms per image and reaches accuracy 28.2 mAP, as accurate as SSD but three times faster \cite{YOLOv3:2018:Redmon}. Hence, in this work, YOLOv3 is adopted to get player bounding boxes.

The YOLOv3 pre-trained model is used to detect the bounding boxes of players in video. However, YOLOv3 marks all of the people in the pictures including non-players. Therefore, based on the information of the badminton court layout, an homography matrix is calculated to estimate the projection 2D coordinate on the court. The midpoint on the lower boundary of a bounding box is considered as the 2D coordinate of the object that is used to filter out non-plyers' bounding boxes. The bounding box coordinates which are outside the court region will be filtered out. Based on the information collected, we can roughly locate and analyze the player's position on the field.

Besides the bounding boxes that can give the position information, the skeletons of players can tell the movement details of players. The bounding boxes of players are enlarged by 0.5 times to enclose a whole image of the player. Then, OpenPose \cite{OpenPose:2018:Cao}, which was developed by Carnegie Mellon University (CMU), is adopted to depict the skeleton on the MPII model for the human near the center of the bounding box. Players on both sides are processed separately. In the MPII model, the human body is sketched in 15 keypoints. Based on the series of skeletons, the player's footwork and physical movements can be understood and analyzed. Furthermore, the playing modes of players in the matches, including defensive playing and offensive playing, could be classified. In the future, a new skeleton model that is composed of the MPII skeleton model as well as an extra keypoint representing the racket will be trained.

In summary, combined with the bounding boxes of players obtained by YOLOv3 and the skeletons of athletes obtained by OpenPose, the movements and footwork can be analysis. Moreover, big data analysis will be introduced to improve the efficiency of players' training and event to adjust playing tactics in their matches.

\section{Implementation}
\label{sec:Implementation}

In this section, we introduce some preliminary results of the CoachAI project. First of all, for training and testing, two matches of Tai Tzu-Ying in 2018 All England Open have been manually labeled, including Tai Tzu-Ying vs. Akane Yamaguchi and Tai Tzu-Ying vs. Chen Yufei. There are about 150,000 frames in the two match videos. In the frame level, the shuttlecock position, bounding boxes of players, and skeletons of players were labeled. In the rally-level, the shuttlecock hit time and ball types were annotated, and the reasons of score or loss were also remark. The rally-level labeling is implemented by the members of the NCTU school badminton team.

\begin{figure}[ht]
    \centering
    \includegraphics[width=0.48\textwidth]{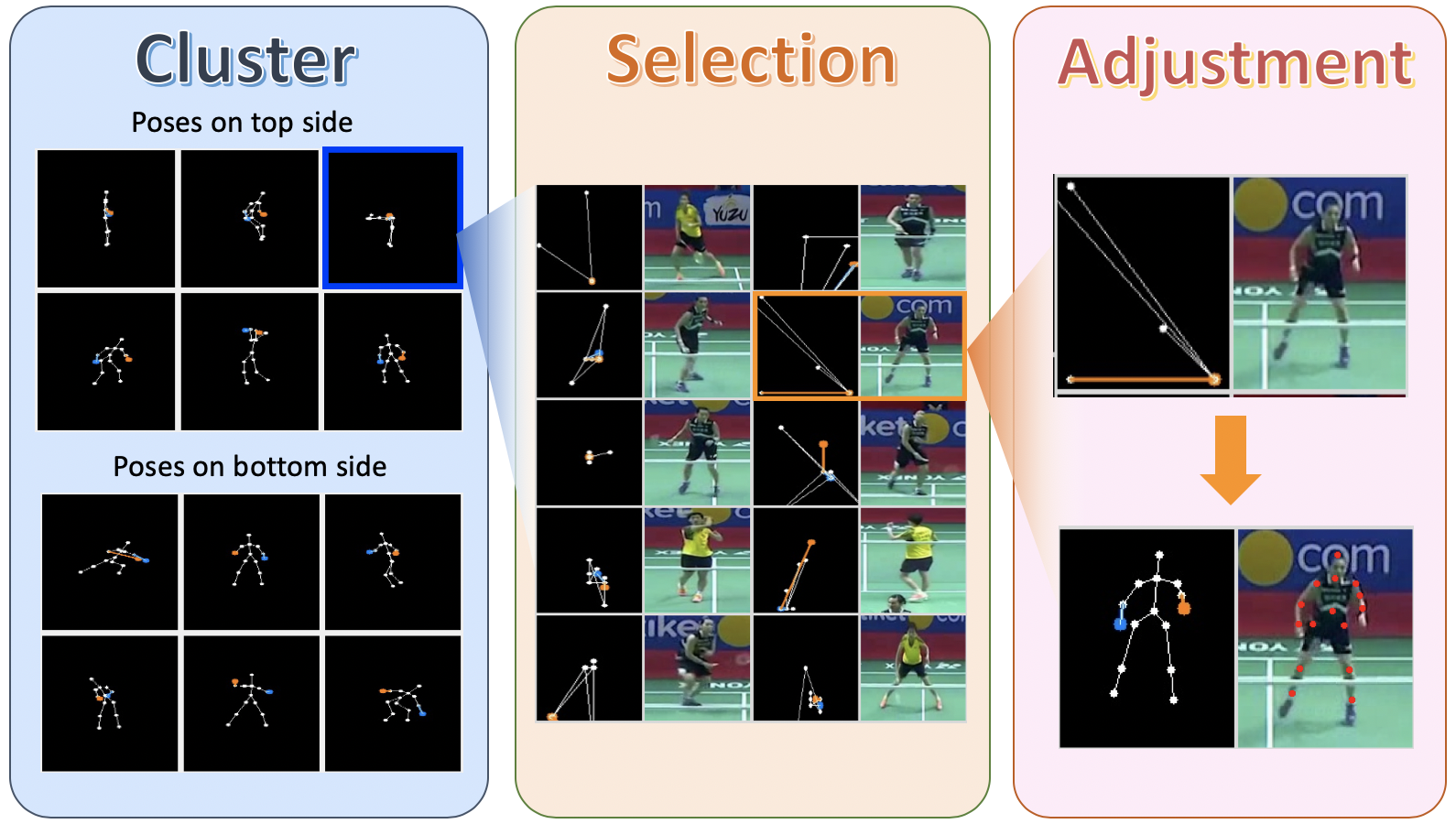}
    \caption{Flow chart for skeleton adjustment.}
    \label{fig:cluster}
\end{figure}

A badminton match can usually last for near an hour. The video may have nearly 100,000 frames. The labeling task is quite laborious. Therefore, a video preprocessing tool helped to mark out non-competition screens, and also provided rough start and finish times of rallies. To increase the efficiency of skeleton labeling, YOLOv3 was used to segment players. After magnifying the edges of the player bounding box by 1.5 times, the segmented image was analyzed by OpenPose package to detect the skeleton in the MPII model. The skeleton data are clustered to identified outliers. Only skeletons considered as outliers are manually labeled. Fig.~\ref{fig:cluster} illustrates the process. The clusters of skeletons detected by the MPII pre-trained model are sketched in the box on the left hand side. If a outlier cluster is selected, many incorrect skeletons with images are displayed. See the pictures in the box at the middle. Only those skeletons will be adjusted. See the pictures in the box on the right hand side. After the skeletons are corrected, the new dataset is used to fine-tune the model. In the future, we will add a keypoint corresponding to the racket neck join in our skeleton model.

\begin{figure}[ht]
    \centering
    \includegraphics[width=0.48\textwidth]{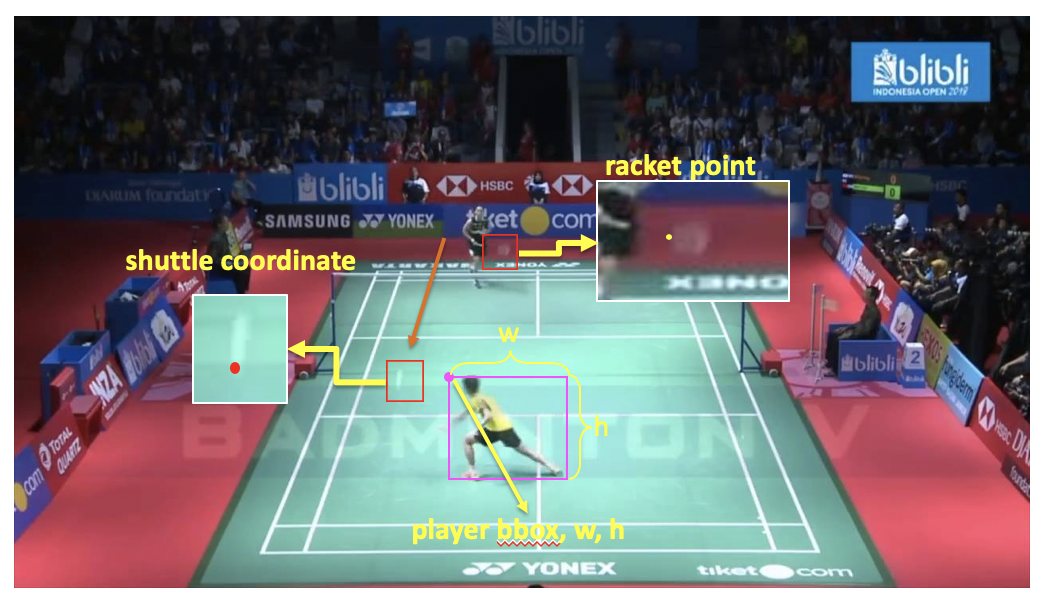}
    \caption{An example of frame-level labeling: This picture is captured from the video of 2018 Indonesia Open Final Match \cite{2018Indonesia}.}
    \label{fig:frame}
\end{figure}

For the labeling of shuttlecock, due to the high moving speed, the shuttlecock images in the video may be blur and with afterimage trace. In such cases, the latest ball image is labeled to be the position of the shuttlecock. This method is the same as the method of labeling tennis proposed in the work of TrackNet \cite{TrackNet:2018:Huang}. The racket keypoint is labeled at the join of the neck and face of the racket. For example, in Fig.~\ref{fig:frame}, the shuttlecock is flying in the direction from the top-side player to the bottom-side player. Therefore, the shuttlecock is labeled at the red dot. In addition, the racket of the top-side player is labeled at the yellow dot.

\begin{figure}[ht]
    \centering
    \includegraphics[width=0.48\textwidth]{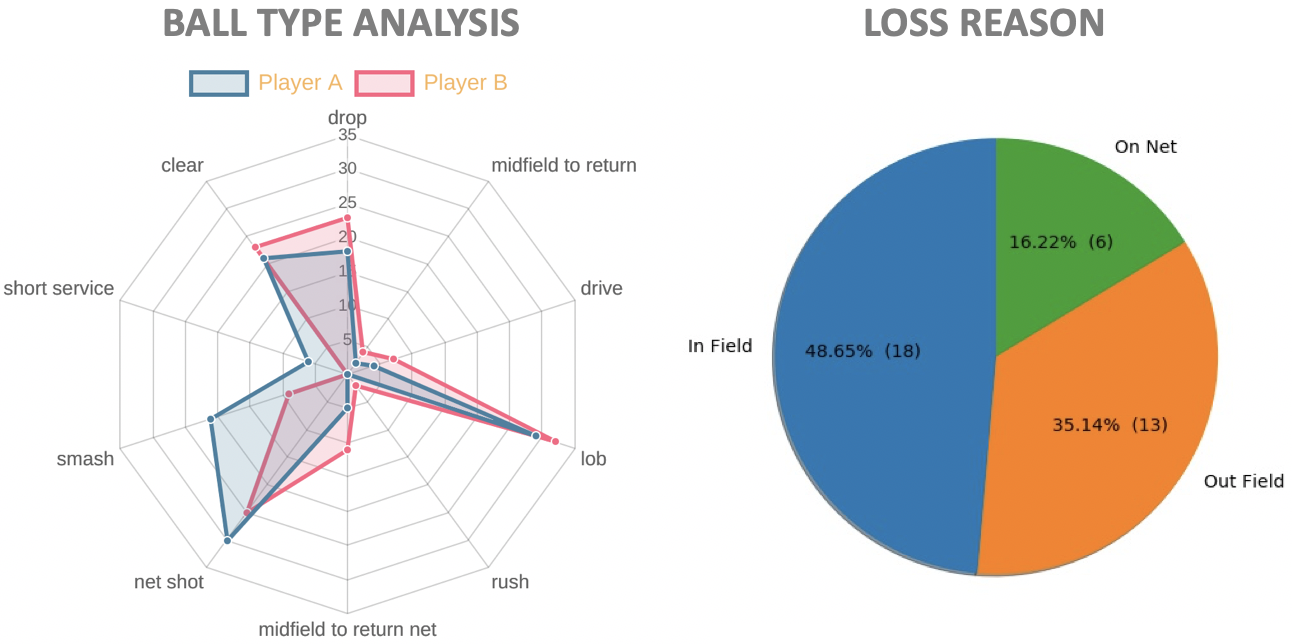}
    \caption{Left: the distribution of the ball type usage. Right: the distribution of the reasons of loss.}
    \label{fig:F_statistic}
\end{figure}

\begin{figure}[ht]
    \centering
    \includegraphics[width=0.48\textwidth]{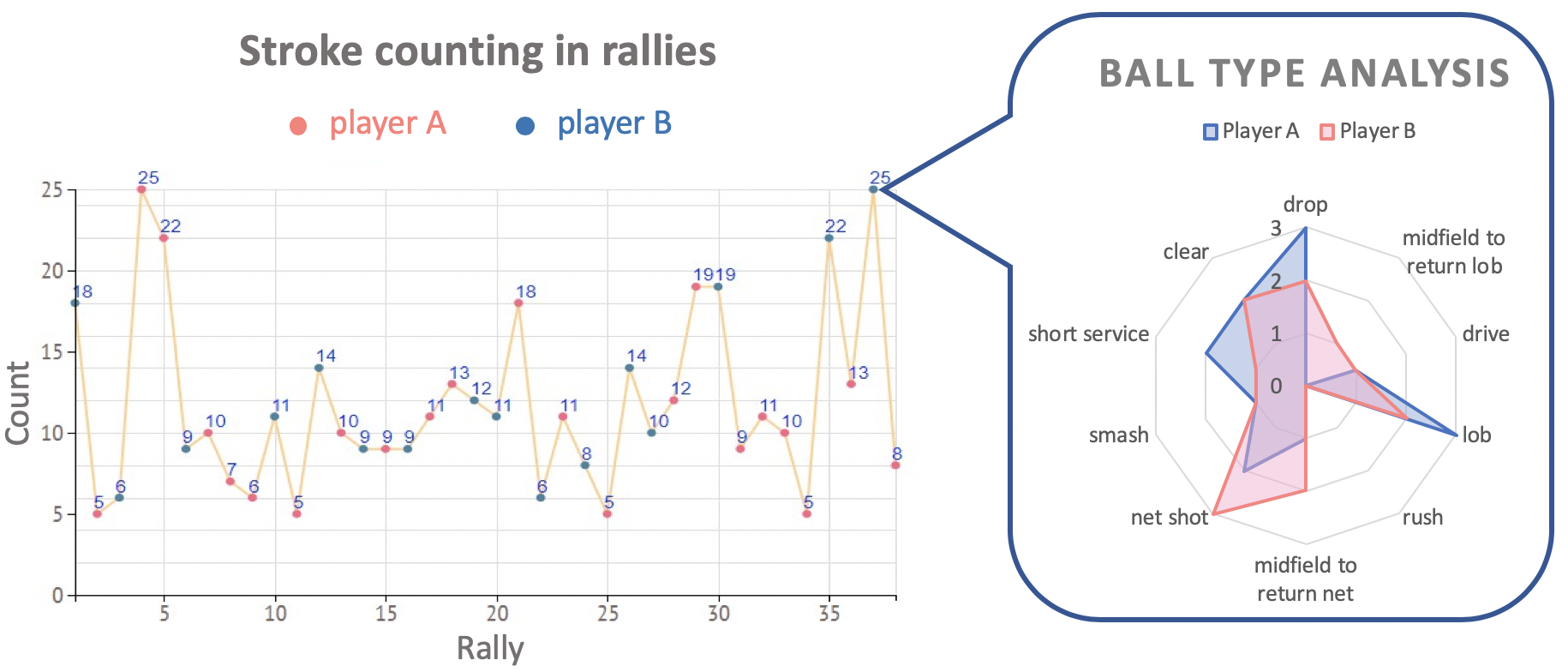}
    \caption{Left: the statistics of stroke count per rally; Right: the distribution of ball type usage in a rally.}
    \label{fig:statistic}
\end{figure}

Tactical analysis and visualization can provide us details and insights of a badminton match. Fig.~\ref{fig:F_statistic} shows an example of the distribution of stroke types and the statistics of loss reasons in a game. It can be observed that the two players use a slightly different stroke type. More rally-level information is illustrated in Fig.~\ref{fig:statistic}. The left sub-figure shows the stroke count for each rally. The color at the vertices denotes the player who scored this rally. As a vertex is clicked, a radar chart like the right sub-figure will pop up to show the statistics of ball type usage in this rally.

\begin{figure}[ht]
    \centering
    \includegraphics[width=0.48\textwidth]{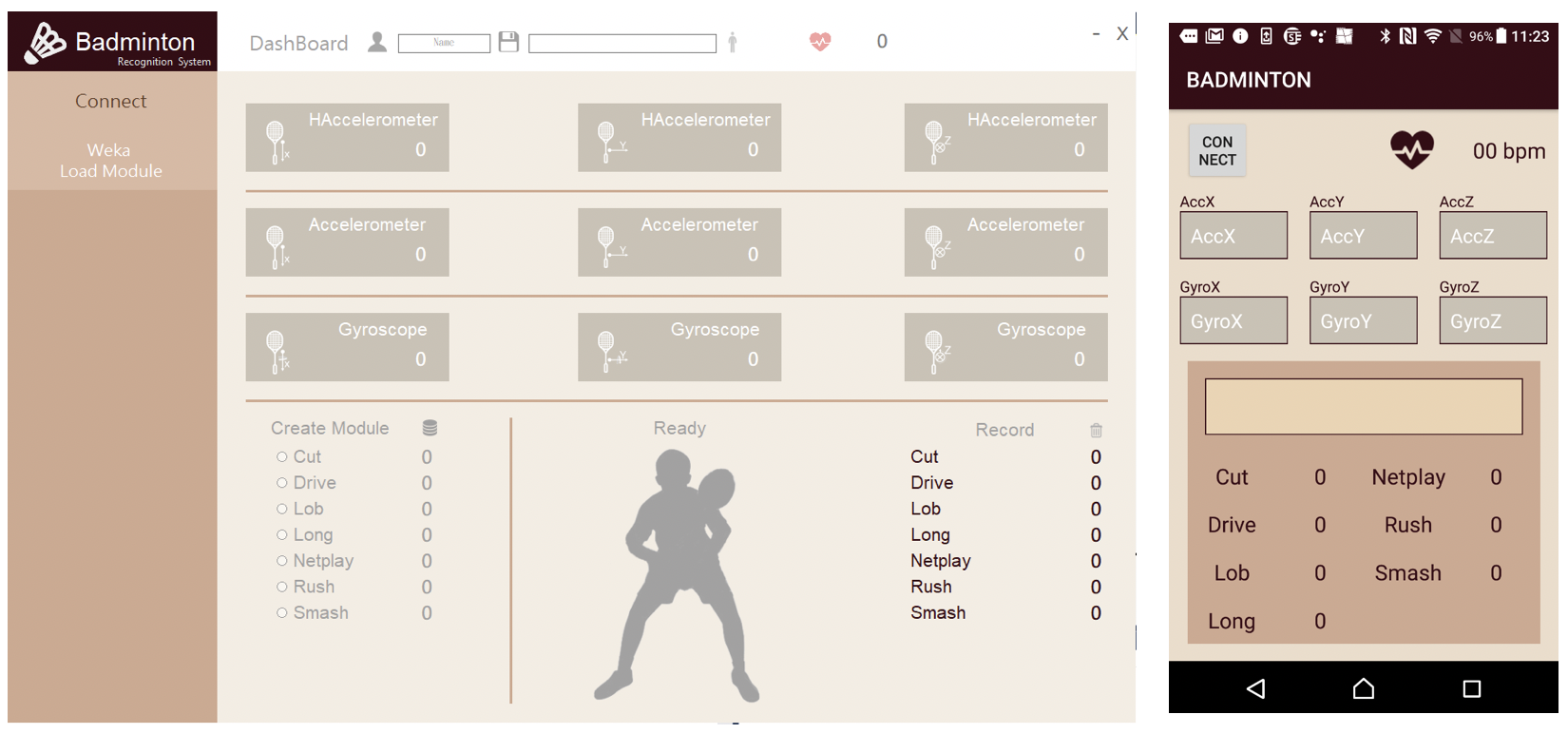}
    \caption{The UI of the smart badminton racket system.}
    \label{fig:interface}
\end{figure}

Fig.~\ref{fig:system} illustrates the system architecture of the smart badminton racket. Fig.~\ref{fig:interface} shows the UI of the smart racket system. When receiving the signal of strokes from the smart racket via Bluetooth, the system classifies the strokes and shows the results on the interface instantly, as illustrated in Fig.~\ref{fig:interface}. Currently, seven common stroke types can be recognized, including cut, drive, lob, long, netplay, rush and smash. The UI has been developed as a mobile app and on web platform.

\section{Conclusion}
\label{sec:Conclusion}

In this paper, a project for badminton competition data collection and tactical analysis has been introduced. Deep learning techniques are adopted to develop video-based competition data collection, including shuttlecock trajectory and players' positions and skeletons. Based on the microscopic data, machine learning techniques are used to develop tactical analysis. Visualization techniques are used to properly and meaningfully represent and display microscopic and macroscopic competition data for easy understanding. Besides, training auxiliary devices including smart badminton rackets and connected serving machines will be developed in order to improve the training efficiency. In the future, besides developing the techniques depicted in the road-map, we would like to develop the capability of 3D data collection and analysis, and extend the research on singles matches to doubles matches.

\balance
\section*{Acknowledgement}
This work of T.-U. \.Ik was supported in part by the Ministry of Science and Technology, Taiwan under grant MOST 107-2627-H-009-001 and MOST 105-2221-E-009-102-MY3.
This work was financially supported by the Center for Open Intelligent Connectivity from The Featured Areas Research Center Program within the framework of the Higher Education Sprout Project by the Ministry of Education (MOE), Taiwan.
We thank Ms. Yu-Ching Kao, Mr. Nien-En Sun and the school badminton team, NCTU for assistance in data labeling, and Advanced Database System Laboratory (ADSL Lab), NCTU in the work of tactical analysis.

\bibliographystyle{IEEEtran}
\bibliography{CoachAI}

\end{document}